# Chat2SPaT: A Large Language Model Based Tool for Automating Traffic Signal Control Plan Management

Yue Wang, Miao Zhou, Guijing Huang, Rui Zhuo, Chao Yi, Zhenliang Ma

*Abstract*—**Pre-timed traffic signal control, commonly used for operating signalized intersections and coordinated arterials, requires tedious manual work for signaling plan creating and updating. When the time-of-day or day-of-week plans are utilized, one intersection is often associated with multiple plans, leading to further repetitive manual plan parameter inputting. To enable a user-friendly traffic signal control plan management process, this study proposes Chat2SPaT, a method to convert users' semi-structured and ambiguous descriptions on the signal control plan to exact signal phase and timing (SPaT) results, which could further be transformed into structured stage-based or ring-based plans to interact with intelligent transportation system (ITS) software and traffic signal controllers. With curated prompts, Chat2SPaT first leverages large language models' (LLMs) capability of understanding users' plan descriptions and reformulate the plan as a combination of phase sequence and phase attribute results in the json format. Based on LLM outputs, python scripts are designed to locate phases in a cycle, address nuances of traffic signal control, and finally assemble the complete traffic signal control plan. Within a chat, the pipeline can be utilized iteratively to conduct further plan editing. Experiments show that Chat2SPaT can generate plans with an accuracy of over 94% for both English and Chinese cases, using a test dataset with over 300 plan descriptions. As the first benchmark for evaluating LLMs' capability of understanding traffic signal control plan descriptions, Chat2SPaT provides an easy-to-use plan management pipeline for traffic practitioners and researchers, serving as a potential new building block for a more accurate and versatile application of LLMs in the field of ITS. The source codes, prompts and test dataset are openly accessible at https://github.com/yuewangits/Chat2SPaT.**

**Index Terms—Large language model, traffic signal control, signal phase and timing, prompt engineering, intelligent transportation system.**

## I. Introduction

Traffic signal control (TSC) is critical for urban transportation systems. At intersections, signal timings are adjusted to effectively alleviate congestion, and intricate phase treatments are utilized to improve traffic safety and deliver priority service. The methods of TSC have evolved from fixed-time control, to rule or optimization based dynamic controls, such as actuated control and self-adaptive control. In recent studies, reinforcement learning (RL) has emerged as a promising new approach for TSC, dynamically evaluating rewards from signal control actions [1].

In practice, however, the fixed-time control remains the most frequently used control strategy for traffic agencies [2], mainly considering the deployment cost and system reliability. Fixed-time control is often operated with time-of-day (TOD) and day-of-week (DOW) plans, and thus traffic engineers need to manage multiple signal control plans for each intersection, the process of which extensive requires manual inputs of various signal phase and timing (SPaT) parameters. Dynamic controls, on the other hand, also require pre-set 'background plans' to define the default phasing scheme and sequence, along with algorithm hyperparameters. For traffic agencies, the management of TSC plans requires time-consuming and tedious human interactions with professional TSC software interface.

The advent of large language models (LLM) offers the chance to automate the TSC plan management with chats, which could potentially reduce human workload and boost TSC efficiency. With their powerful reasoning and natural language processing (NLP) capabilities, LLMs are able to transform users' semi-structured and potentially ambiguous descriptions to structured data, and streamline the interactions between human and domain specific software. To evaluate LLMs' performance in the field of traffic engineering, extensive efforts have been made in recent studies, using fine-tuning or few-shot prompting techniques. The topics of TSC, transportation safety and autonomous driving have drawn the most interest from researchers [3], [4]. Focusing on the field of TSC, this paper categorizes LLM's application in existing studies into three groups.

**Functional level**. In the process of TSC designing and management, various types of peripheral tasks are conducted by traffic engineers, including multimodal data processing, road network analysis, database query, report generation, and traffic simulation. To improve efficiency and accuracy, LLMs are utilized to aid traffic engineers in these tasks. In [5], a fine-tuned model, TransGPT, is proposed for the reasoning with multi-modal transportation knowledge, and is validated by QA tests. In [6], platform CityBench is proposed for diverse urban transportation tasks, such as GIS (geographic information system) related semantic understanding and mobility prediction. The tool ChatSUMO is proposed in [7] to enable simulation scenario generation through chats. LLMs are used to convert user descriptions to keywords, with which scripts are designed to generate simulation networks.

*Corresponding author: Miao Zhou.*

Yue Wang, Miao Zhou, Rui Zhuo and Chao Yi are with Zhejiang Dahua Technology Company Ltd., Hangzhou 310053, China (e-mail: wang.yue3@northeastern.edu; dahuaitslab@163.com; 13180839671@163.com; yichaov5@163.com).

Guijing Huang is with Hangzhou AliCloud Apsara Information Technology Co., Ltd., Hangzhou 310000, China (huanggjcs@126.com).

Zhenliang Ma is with the KTH Royal Institute of Technology, 100 44 Stockholm, Sweden (e-mail: zhenliang.ma21@gmail.com).

**Strategic level**. Leveraging LLMs' skills of reasoning and task planning, traffic management frameworks are developed, combining LLMs and transportation functions. Prompts are designed for each transportation function [8], [9], [10], outlining function's name, input, output, example, priority, etc. These prompts are further included in the system prompt, to help LLMs select the most relevant tools, for perception queries or decision making tasks. Such frameworks could understand fuzzy requests or inquiries, clearly demonstrate reasoning steps and interactively aid traffic engineers in tasks such as traffic analysis, data processing, decision making and performance evaluation.

**Tactic level**. LLMs are utilized as agents to control traffic signals, complementing traditional TSC methods based on traffic engineering and reinforcement learning methods. In [11], an LLM-based workflow is proposed to generate signal control plans for urban arterial roads. LLMs are instructed to analyze arterial road networks, understand the theory of coordinated control, and calculate offsets among intersections accordingly. Yet, the focus of the study, one-way coordination, is a trivial problem, and it does not consider bidirectional green wave coordination with LLMs. Similarly, step-by-step calculation of Webster's method is broken down for LLMs in [12]. In the same study, LLM is also utilized to aid reinforcement learning for TSC, which outperforms fixed-time and actuated control models. However, the study only tests simple phasing schemes, with only 2 actions (either green or red) defined in the action space for each approach of the intersection, which means that no phasing scheme with protected left-turns is allowed in the signal optimization. LLMs are also used to perceive traffic conditions and take over dynamic traffic signal control. In [13], LightGPT, a fine-tuned model, is prompted to select the signal for the lanes with the highest congestion. The model uses vehicle counts as congestion indicators, and outperforms multiple state-of-the-art (SOTA) reinforcement learning based models. However, using a similar idea, authors in [6] fail to replicate such results, and find that the performance of LLMs are even far worse than traditional methods such as Max-Pressure. To leverage LLMs' reasoning capabilities in TSC, some research focus on the application of LLMs in user-defined scenarios. In [10], LLMs are instructed to address cases of roadblock incident, emergency vehicles, and sensor outage. LLM based methods significantly outperform traditional methods, for which such special scenarios are not considered.

In traditional methods, an effective TSC often requires complex mathematical models, such as optimization, considering trade-offs among movements, traffic condition evolvement based on traffic flow theory and traffic safety policies. A practical TSC model may not be easily represented by simple rules of finding the approach with the most queued vehicles or the largest detector occupancy, as tested in [6], [9], [10] and [13]. In other words, TSC is not a purely text and reasoning based task by nature, and replacing traditional TSC algorithms solely with language models at their current stage may not be the best option. Instead, using LLMs to aid traffic engineers in TSC strategically seems to be a more practical direction in the near future. In such systems, LLMs are leveraged to understand context and users' requests, interactively provide decision-making support with reasoning, and integrate building blocks of TSC through task planning.

In this study, targeting at the management of traffic signal control plans, one of the most fundamental building blocks for TSC, we aim to improve traffic engineers' workflow by automating the process of signal plan creating and editing using chats. An LLM-powered tool, Chat2SPaT, is proposed, which is capable of converting users' descriptions on TSC plans to executable SPaT results. Prompts are curated to extract phasing sequence and phase attributes from semi-structured chats, and scripts are designed to integrate nuances of traffic engineering into the assembly of TSC plans. The tool covers common description styles and phasing schemes, including stage and ring based structures. Chat2SPaT is extensively tested with a dataset containing 306 plan description cases, in both English and Chinese. As the first benchmark to evaluate LLMs' capability of understanding complex traffic signal control plan descriptions, Chat2SPaT demonstrates practical value for the industry, and serves as a potential new building block for a more accurate and versatile application of LLMs in the field of ITS.

The major contributions of this study are summarized as follows:

1) The study proposes Chat2SPaT, an LLM powered tool to enable automated chat based TSC plan management. This innovation allows for TSC plan creating and editing based on user descriptions, relieving traffic engineers of tedious manual SPaT parameter inputting and boosting efficiency for traffic agencies.

2) An open-access test dataset with over 300 plan descriptions in both English and Chinese is created, covering common TSC plan schemes and special phase treatments. Through extensive tests, the results show that the proposed methodology could understand a variety of description styles, and accurately generate and validate TSC plans.

3) In Chat2SPaT, most traffic engineering nuances and numerical calculations are handled by TSC specific scripts, which minimizes the requirement for LLMs' capability in complex mathematical reasoning. Moreover, fault tolerance functions are developed, targeting at typical errors in LLM outputs. These features make Chat2SPaT robust against LLM size, increasing the tool's feasibility to apply in practice, where large-sized LLMs are not available.

This paper unfolds as follows: Section II gives an overview of the proposed methodology. The two main components of Chat2SPaT, prompt engineering and script-based plan assembly, are introduced in Section III and Section IV, respectively. Section V presents experimental results of the tool's performance in traffic signal control plan management. Section VI concludes the paper.

## II. Methodology Overview

In this paper, Chat2SPaT, an iterative two-step workflow methodology is proposed for an efficient and accurate chat-based traffic signal control management. Chat2SPaT leverages LLMs to understand semi-structured and ambiguous plan descriptions, and then uses rule-based algorithms to generate complete SPaT results, with domain knowledge integrated in both prompts and scripts.

First, prompts are designed for LLMs to accurately understand users' description of plan creating and editing. The design of prompts takes into consideration various styles of plan descriptions in terms of temporal relationships among phases within the cycle, such as stages, rings, and phase overlapping. To avoid LLMs' incorrect reasoning when facing complex plans and potential ambiguity in the descriptions, the prompts clearly instruct LLMs to record special phase treatments as phase attributes with a key-value format, such as permissive left-turns and exclusive pedestrian crossing phases. LLMs are then instructed to provide three results as a pre-defined json object, including phase sequence, phase attributes, and cycle length.

Second, based on the results from LLMs, python scripts are designed to locate phases in a cycle and formulate the complete plan. Temporal information of movements is extracted by categorizing all phases into three groups: major phases (phases listed in stage-based or ring-based structures), overlapped phases (running concurrently with a major phase) and standalone phases (phases described with their own start and end times). Rule-based algorithms are then developed to integrate phase attribute results into each phase to implement users' special phase treatments and deal with nuances of traffic signal control. Finally, the plan results are provided as movement-level SPaT information. With contexts retained in the chat, the proposed workflow could be utilized iteratively to enable plan modification, so that users could generate a plan step by step or make editing on a template. Moreover, the tool validates plan results and reminds users to apply further modification, if the generated plan is invalid due to conflicted movements or short pedestrian walk intervals.

## III. Prompt Design

Considering that users may describe the plan in different styles, with potentially ambiguous and incomplete specifications, prompts are designed to help LLMs understand users' descriptions of plan creating and editing. This section introduces the detailed prompt design to convert users' plan descriptions to structured results, and then shows how the variations of different movements and plan modifications can be handled under this prompt design.

The prompts first explain various plan structures, including common stage and ring based schemes and terms such as split phasing, exclusive pedestrian phase, and 'permissive only' phasing (two opposing approaches timed concurrently). With examples, LLMs are instructed to record the 'major phases' and their timing parameters in each type of plan structure.

The prompts then instruct LLMs to record overlapped phases and standalone phases. Overlapped phases are added by assigning a major phase occurrence in the plan structure as the 'parent phase', whereas standalone phases are added by directly recording its timing parameter attributes without referring to other phases. For each phase occurrence in the plan, various attributes are recorded in a key-value format, to enable special phase treatments and complex plans.

Further, the prompts list possible modification descriptions on phase attributes and plan structures, and clearly show LLMs the steps to apply the modifications on the current results.

Finally, the prompts instruct LLMs to generate three json-style results: phase sequence result, phase attribute result and cycle length result, which will be used to generate exact SPaT results.

### A. Overall Prompt Design

In Chat2SPaT, a user-interface is provided for users to see the available movements and the phase coding, with an example shown in Fig. 1. The available movements should be defined based on the signal heads and layout of the intersection. In this study, to fully test the model's capability of generating and editing various types of signal control plans, it is assumed that all the movements in Fig. 1 are available for the intersection, including 16 vehicular movements and 12 pedestrian crossing movements.

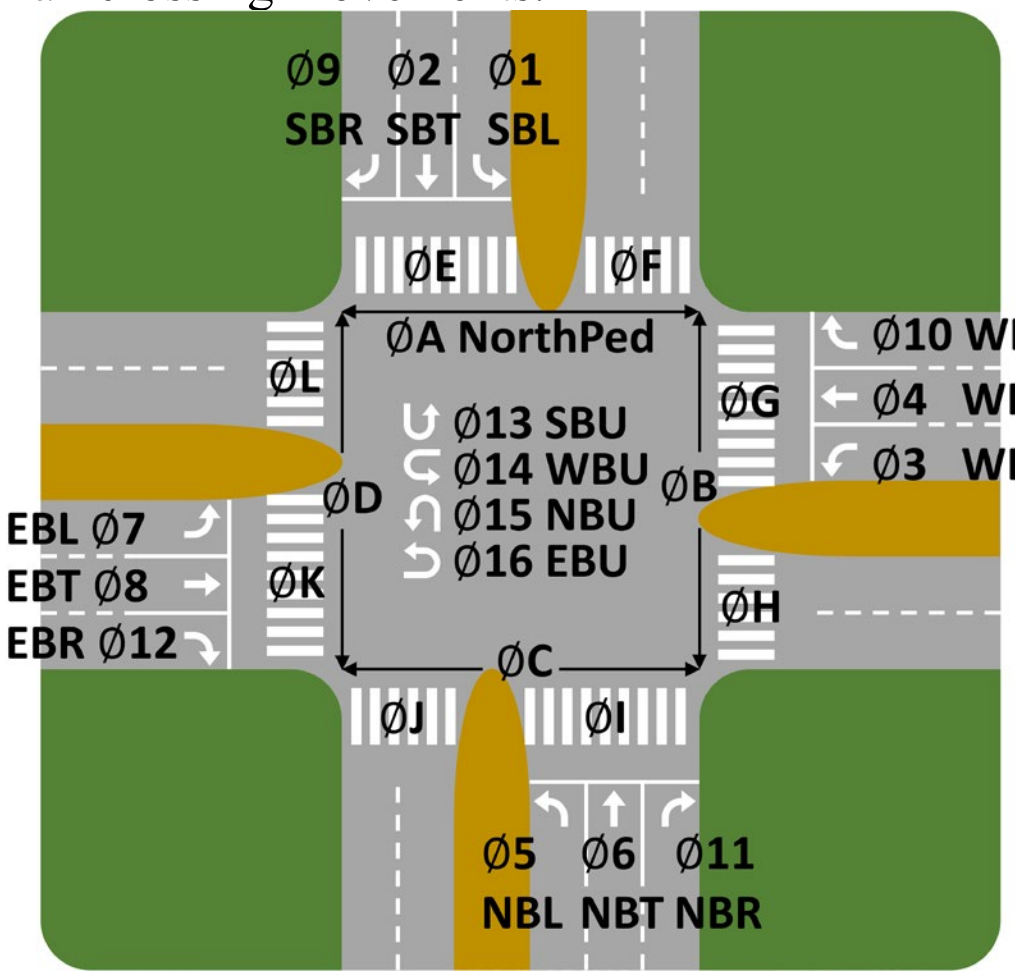

**Fig. 1.** Movement phase coding for a typical four-legged intersection.

Using formal phase names, or the abbreviations / coding in Fig. 1, users specify the sequence and duration of phases in the traffic signal control plan. It is common practice that traffic engineers design signal control plans with a stage or ring-based plan structure, such as the dual-ring structure defined following NEMA (National Electronics Manufacturing Association) convention. In this paper, phases listed in such plan structure are referred to as 'major phases', which are typically through and left-turn movements, related to the most severe cross conflicts at intersections [14]. Similar to the design process of signal control plans, a user may selectively start the plan description with 'major phases' in a stage or ring-based plan structure, and then add other phases in a second step. In Fig. 2, the red arrows and associated texts illustrate four possible styles of plan structure description, which are based on rings, stages, or a combination of them.

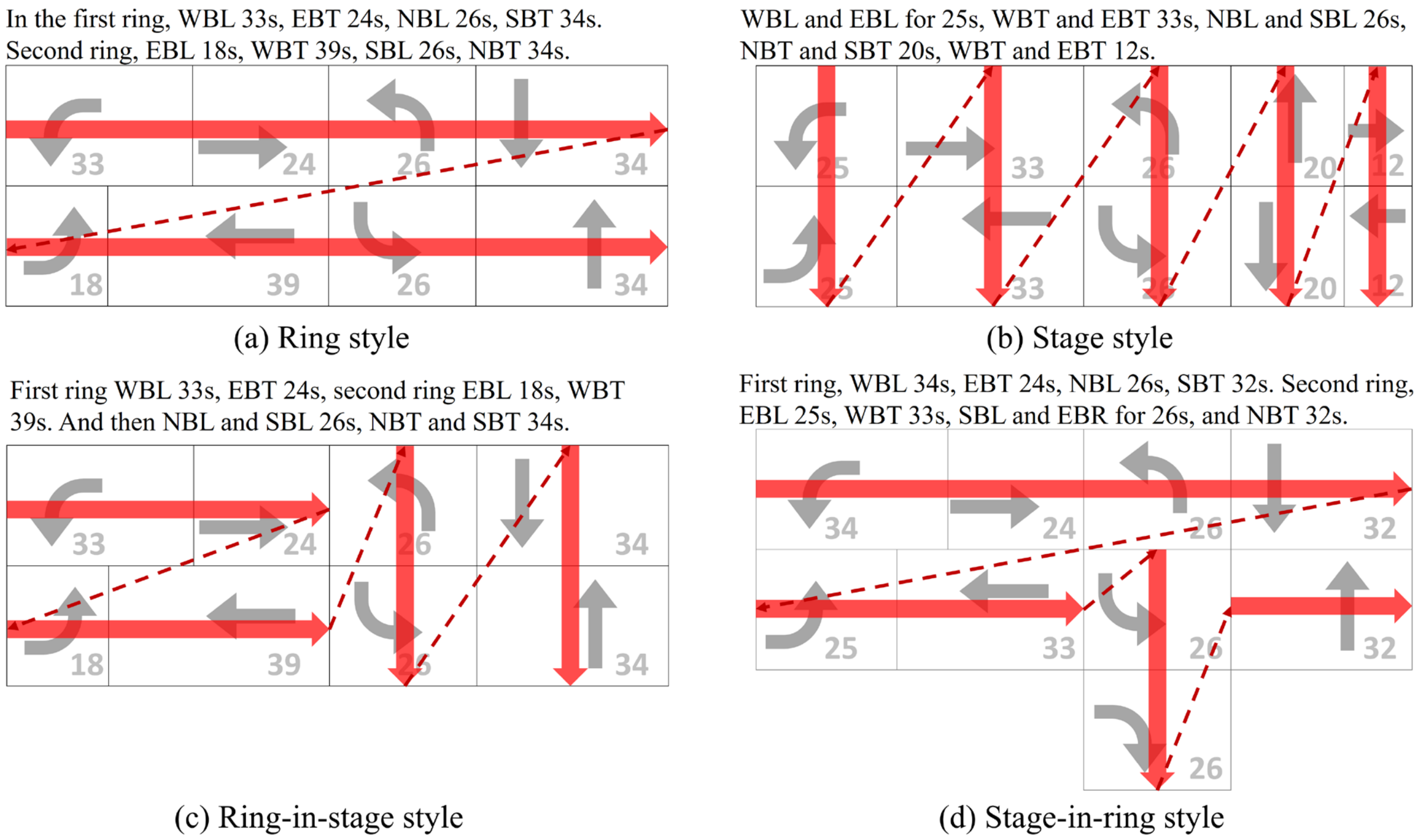

**Fig. 2.** Various description styles for phase sequence.

Based on users' description, LLMs are instructed to record the plan structure in the phase sequence result. The phase sequence result is written as a list of stage or ring objects, with key 'stageStyle' or 'ringStyle' to label the structure style. For each 'stageStyle' ('ringStyle') object, the value is a nested list, with phases in the same stage (ring) recorded sequentially in the inner lists. For each phase in the phase sequence result, it is assumed that the time length provided by the user is the split duration, and is recorded as the value for the key 'split' in the phase object. Otherwise, if the time length is claimed to be the green time duration, the key of 'greenTime' would be used instead.

Using the major phases in the plan structure defined above as parent phases, some other phases, which are typically pedestrian crossing movements and right turn movements, could be added into the plan by phase overlapping. In this paper, such phases are referred to as 'overlapped phases', which are recorded in the phase attribute result.

When an overlapped phase is specified to overlap a major phase, the major phase name is recorded with the attribute 'parentPhase'. With this overlapping relationship, overlapped phases could be located in the cycle, and thus, LLMs do not need to calculate the timing of overlapped phases.

In users' descriptions, phase overlapping may not always be a precise one-to-one mapping. And prompts are designed to instruct LLMs to handle potential ambiguity or incompleteness in phase overlapping.

First, the major phase to overlap may have more than one occurrences within a cycle, either because of stage-based description or phase re-service. For example, in Fig. 2a, users may describe movement WBT in two stages, i.e. a 15-second stage with WBL and a 24-second stage with EBT. If the user further specifies that ØA (pedestrian crossing movement on the northern approach) overlaps WBT, it is assumed that all occurrences of WBT are overlapped. In the case of phase re-service, however, users may specify which occurrence of the major phase to overlap. For example, in Fig. 2b, users may specify ØA to overlap only the first occurrence of WBT, as the second WBT is too short to cover the minimum walk interval plus clearance time of the pedestrian crossing movement. In this case, in addition to recording the name of the major phase, LLMs are also instructed to record the specified number of the occurrence to overlap, with attribute 'overlapNum'.

Second, a phase may overlap multiple major phases. Using right-turn movements as an example, based on traffic volume data, the user may decide that the right-turn movement EBR is compatible with both EBT and NBL, and specify that EBR overlaps these two major phases in the plan description. In this case, two phase objects are created for EBR, with attribute 'parentPhase' being EBT and NBL, respectively. If these two EBR phase occurrences are consecutive in the cycle and form a single EBR phase, they will be merged by python scripts in the next section.

If the start time and end time (or equivalently, start time and split) of a phase are provided by the user, then the phase is treated as a 'standalone phase'. For a standalone phase, the provided start time and end time would be prioritized to be used to directly locate the phase in the cycle, despite the phase's position in the phase sequence result or overlapping information. Similar to overlapped phases, standalone phases are also recorded directly in the phase attribute result.

For each occurrence of all phases, a phase object is created in the phase attribute results, as a collector of various phase attributes. The start time, end time and duration of the split or green time of a phase, if provided by the user, are recorded as phase attributes. Inter-green parameters, such as yellow time, red/amber time and all-red time are also adjustable as phase attributes. Users could further apply late start or early cut-off control parameters, to enable special phase treatments such as leading pedestrian interval (LPI). A phase could also be set as permissive or prohibited to construct more complex plans, as will be introduced in part B of this section.

If the end time of a phase is specified implicitly by the user, for instance, "NBT runs till the end of the cycle", a place holder 'cycleLength' will be assigned as the value for phase attribute 'endTime'. In this case, the value of cycle length could be derived from the major phases in the plan structure. Alternatively, if cycle length is provided in the description, LLMs are instructed to record it in the cycle length result.

In Fig. 3, an example of plan description is shown, with specification of cycle length, major phases, overlapped phases, standalone phases, and various phase attributes. Below the description, the phase sequence result, phase attribute result and cycle length result are recorded as "result1", "result2" and "result3" in the LLM outputs, respectively.

First ring NBL 21s, SBT 44s, second ring SBL 39s, NBT 26s. And then WBL and EBL green for 17s, WBT and EBT green time 22s. All phases 3 seconds of green flash. NBR overlaps WBL. WBR runs from 88s to 21s. Add default pedestrian phases.

```json
{
  "result1": [
    {
      "ringStyle": [
        [{"NBL": {"split": 21}}, {"SBT": {"split": 44}}],
        [{"SBL": {"split": 39}}, {"NBT": {"split": 26}}]
      ]
    },
    {
      "stageStyle": [
        [{"WBL": {"greenTime": 17}}, {"EBL": {"greenTime": 17}}],
        [{"WBT": {"greenTime": 22}}, {"EBT": {"greenTime": 22}}]
      ]
    }
  ],
  "result2": [
    {"NBL": {"phaseOrder": 1, "greenFlash": 3}},
    {"SBT": {"phaseOrder": 1, "greenFlash": 3}},
    {"SBL": {"phaseOrder": 1, "greenFlash": 3}},
    {"NBT": {"phaseOrder": 1, "greenFlash": 3}},
    {"WBL": {"phaseOrder": 1, "greenFlash": 3}},
    {"EBL": {"phaseOrder": 1, "greenFlash": 3}},
    {"WBT": {"phaseOrder": 1, "greenFlash": 3}},
    {"EBT": {"phaseOrder": 1, "greenFlash": 3}},
    {"NBR": {"phaseOrder": 1, "greenFlash": 3, "parentPhase": "WBL", "overlapNum": 0}},
    {"WBR": {"phaseOrder": 1, "greenFlash": 3, "startTime": 88, "endTime": 21}},
    {"NorthPed": {"phaseOrder": 1, "parentPhase": "default"}},
    {"SouthPed": {"phaseOrder": 1, "parentPhase": "default"}},
    {"EastPed": {"phaseOrder": 1, "parentPhase": "default"}},
    {"WestPed": {"phaseOrder": 1, "parentPhase": "default"}}
  ],
  "result3": null
}
```

**Fig. 3.** Example plan description and LLM outputs.

### *B. Left Turn Movements*

There are five common options for the left-turn phasing: permissive only, protected only, protected-permissive, split phasing, and prohibited [15]. To serve these phasing options, in the prompts, a left-turn phase is assumed to be protected when mentioned by the user, and could be specified as permissive or prohibited with attributes 'isPermissive' and 'isProhibited', respectively.

For protected-permissive left-turns, which are typically operated with five-section signal heads (also known as 'dog house'), users may add a normal protected left-turn and a permissive left-turn consecutively. For simplicity, in this study, lights-off are applied for permissive left-turns, despite that in some areas in the US, a flashing yellow arrow (FYA) is utilized for permissive left-turns, which is found to reduce left-turn related crashes [16].

When split phasing is applied, LLMs are instructed to create a through movement and a protected left-turn movement for the direction. Unless specified by users, U-turn movements, which typically run concurrently with the associated left-turn movements, are ignored by default.

### *C. Right Turns Movements*

By default, when a right-turn is not mentioned by the user, it is assumed to be permissive for the entire cycle. This means no signal heads or lights off for this right-turn movement, and it will be ignored in the generated plan.

In [17], protected right-turns are recommended to overlap the cross-street's protected left turn. In practice, however, protected right-turns may also be timed in other stages, based on the volume and potential crash risk of each movement. Similar to left-turns, permissive right-turns may also be jointly used with protected right-turns or red lights, to enable a more flexible control.

In the description, right-turn movements could be added by overlapping major phases. Alternatively, for a right-turn phase that covers multiple stages or has late start / early cut-off, it is more convenient to treat it as a standalone phase. Fig. 3 shows two examples of right-turn movement descriptions.

### *D. Pedestrian Movements*

For four-legged intersections, the typical strategy for the operation of pedestrian movements is to run them concurrently with the adjacent through movement phases [15]. Therefore, if the start time and end time of a pedestrian phase are not explicitly provided in the description and cannot be derived from stage or ring-based plan structures, it is treated as an overlapped phase, running concurrently with the adjacent through movement. For example, in Fig. 1, the pedestrian movement on the northern approach (ØA) runs concurrently with WBT (Ø4) by default. In the prompts, the default through movement for each pedestrian crossing could be provided for LLMs to refer to. However, in experiments, it is found that LLMs sometimes ignore the clearly stated default through movements, but assign incorrect parent phases for pedestrian crossing phases. For example, LLMs may select NBT as the parent phase for the northern pedestrian crossing (ØA), probably because these two movements are both associated with the direction of 'north'. Therefore, to avoid such incorrect reasoning, a more effective solution is to instruct LLMs to write down 'default' for the value of key 'parentPhase' as a place holder, and assign the correct default through movement phases later with scripts.

For large intersections, two-stage pedestrian crossing phases are used to improve safety and overall intersection capacity. This is especially effective for intersections of a major road and a minor road, where the time needed for a single-stage crossing is considerably greater than the demand of the concurrent traffic on minor road [18]. When a two-stage pedestrian crossing is specified in the description, LLMs are instructed to create two separate movements for the associated direction. For example, in Fig. 1, the two-stage pedestrian crossing movements on the northern approach are North Ped A (ØE) and North Ped B (ØF), with letter A denoting the side of entering traffic and B denoting exiting traffic. In a typical four-stage control plan (as shown in Fig. 2d), a two-stage pedestrian crossing movement may overlap with two consecutive stages in the plan. For example, ∅E overlaps with WBT and EBL, and ∅F overlaps with WBT and SBL. Similar to one-stage pedestrian crossing, LLMs are also instructed to use the place holder 'default', when the parent phases for two-stage pedestrian crossing movements are not specified by the user.

If exclusive pedestrian phase (also known as a pedestrian scramble or Barnes Dance) is used, LLMs are instructed to add a stage with a phase named 'AllPed' in the phase sequence result and phase attribute result, and the 'AllPed' phase would be broken down into individual pedestrian crossing phases by scripts. Compared with instructing LLMs to directly write down all individual pedestrian crossing movements in the phase sequence result, using the place holder 'AllPed' makes it easier for LLMs to deal with pedestrian crossing movements, as it reduces the chance of hallucination of adding a pedestrian stage for overlapped pedestrian crossing phases when exclusive pedestrian phase is not applied.

### *E. Plan Editing*

With context retained in the chat, LLMs could continue the interaction [7]. Based on the current plan, users may make further modifications. Possible plan editing options include changing the duration of phase split (or green time), adjusting phase attributes, phase addition or removal, alternating plan scheme for a stage or direction, and changing phase overlapping. LLMs are instructed to modify the associated phase sequence and attribute results in the output to implement these editing. Optionally, with descriptions of common signal control plans provided as templates, users could select a plan, append plan editing descriptions to the template description, and use the combined descriptions as inputs for LLMs to construct new plans.

## IV. Plan Assembly

This section assembles the signal control plan by locating each phase in the cycle. Based on the phase sequence and phase attribute results by LLMs, python scripts are designed to conduct data cleansing and streamline plan assembly, integrating nuances of traffic signal control into the specified control parameters. Fault-tolerance functions are designed to address potential errors in LLMs' outputs. The generated plan is validated, and users are warned if conflicted movements are timed simultaneously.

### *A. Data Cleansing*

Data cleansing is conducted on the outputs of LLMs, to standardize phase names, replace place holders and correct potential erroneous records in the results.

If phase codes in Fig. 1 are used to represent phases in LLMs' outputs, they would be replaced by the associated standard abbreviated phase names. In the prompts, place holder 'default' is assigned as the default the value of attribute 'parentPhase' for pedestrian crossing movements. In this step, the place holder would be replaced by the name of the actual parent phase, i.e., the adjacent through movement. When exclusive pedestrian phase is used, the phase object named 'AllPed' in the LLMs' outputs would be replaced by the associated pedestrian crossing phase objects of each direction, which is a process of breaking down the 'AllPed' signal group into individual signal heads.

For an accurate plan generation by scripts, it is critical to ensure value correctness and data format consistency in the results of LLMs. However, due to ambiguity in users' plan descriptions and model hallucination, it is found that LLMs may generate some erroneous records, even with curated prompts targeted at such errors. Therefore, in the step of plan assembly, it is necessary to design fault tolerance functions to address potential errors by LLMs.

Data formatting errors could be corrected at the step of data cleansing. For example, LLMs are instructed to record ring-based plan structure as a nested list, i.e., list of phases inside list of rings. If there is only one ring in the plan structure, LLMs may neglect the list of rings and record only a list of phases. In this case, the recorded ring-based structure would be reformatted as a nested list. For stage based plans, LLMs are instructed to record the stages in the phase sequence result with key 'stageStyle', to distinguish from ring structures. But LLMs may simply write down a list of phases as a natural representation of 'stages', and ignore the requirement to record the key 'stageStyle'. For this case, a fault tolerance function is designed to assign the key 'stageStyle' for such unlabeled lists and reformat the phase sequence result.

### *B. Phase Timing Calculation*

The core for plan assembly is to derive the start and end times of each phase, by integrating the phase sequence and phase attribute results.

The timing of major phases could be calculated based on the stages or rings in the plan structure. For ring-based structures, the start times and end times are calculated sequentially by adding the split duration of each phase in the ring. Similarly, for stage-based structures, the phases in each stage share the same start time, and the end time is calculated using split duration, as shown in (1). In each stage of the plan, the largest phase end time is selected as the stage's end time, and is used as the start time for the next stage.

$$T_{end}^{S} = \begin{cases} C & \text{if } T_{start}^{S} + S = C \\ \text{mod}(T_{start}^{S} + S, C) & \text{o.w.} \end{cases} \quad (1)$$

where $T_{start}^{S}$ and $T_{end}^{S}$ are the start and end times of the split, $S$ is the split's duration, and $C$ is cycle length.

Alternatively, if the start time and end time (or equivalently, start time and split) of a phase are explicitly provided by the user, the phase is treated as a standalone phase, and its location in the cycle can be determined without referring to the plan structure. If a place holder 'cycleLength' is recorded for a phase's 'endTime' attribute, it is replaced by the value of the actual cycle length, which is derived as the largest end time of all major phases or the value in the cycle length result by LLMs, if provided.

Instead of focusing on phase splits, users may provide attributes related to the green time, such as green time duration, start and end times of green, as shown by the east and west bound movements in the phase sequence result in Fig. 3. In this case, these green time related attributes should be converted to split timing parameters.

For simplicity, different from the stage and inter-stage convention in the UK [19], this paper stipulates that the inter-green interval belongs to the split of the associated phase. Hence, split duration is calculated with green time duration and inter-green parameters, as shown in (2).

$$S = LS + RA + G + Y + AR + EC \quad (2)$$

where $LS$ is late start, $RA$ is red/Amber, $G$ is green time duration, $Y$ is yellow, $AR$ is all-red, and $EC$ is early cut-off.

Using the start time of green $T^G_{start}$ and end time of green $T^G_{end}$ , $T^S_{start}$ and $T^S_{end}$ are calculated as in (3) and (4), respectively.

$$T^S_{start} = T^G_{start} - LS - RA \quad (3)$$

$$T^S_{end} = T^G_{end} + Y + AR + EC \quad (4)$$

### *C. Phase Occurrence Merging*

In the plan descriptions, a phase may be added as multiple occurrences in adjacent stages within a cycle, and these 'connected' occurrences should be merged to formulate one complete phase. Two phase occurrences are defined as connected if they are both protected (or both permissive) movements and their time intervals are adjacent (or have an overlap), with consideration of late start and early cut-off attributes. The collection of all the connected occurrences of a phase, denoted as $\boldsymbol{O}$, is generated using Depth First Search (DFS) algorithm. For any phase occurrence $k \in \boldsymbol{O}$, if its start time $t^{start}_k$ is larger than its end time $t^{end}_k$, which means this phase occurrence extends to the next cycle, then it is replaced by two subintervals, $[t^{start}_k, cycleLength]$ and $[0,\ t^{end}_k]$.

Within the cycle, each second $t$ is compared against the phase occurrences in $\boldsymbol{O}$. For any phase occurrence $j \in \boldsymbol{O}$, if $t^{start}_j \leq t \leq t^{end}_j$, $t$ is labeled as part of the merged phase. The combination of all the labelled time points would form the time interval for the merged phase. As all the occurrences in $\boldsymbol{O}$ are 'connected', it is guaranteed that only one time interval would be formed for the phase.

### *D. Phase Overlapping*

For any phase not located in the cycle as a major phase or a standalone phase, if its 'parentPhase' is specified in the phase attribute result, the phase is added as an overlapped phase. The occurrence of the associated major phase (parent phase) is selected, by matching its 'phaseOrder' attribute with the overlapped phase's 'overlapNum' attribute. The start time and end time of the parent phase occurrence are assigned to the overlapped phase, as they share the same split. The green flash duration and inter-green parameters are also assigned to the overlapped phase, such as red/amber, yellow and all-red. To enable flexible phase treatments within the shared split, the overlapped phase and the parent phase could have different late start and early cut off parameters.

If the overlapped phase is a pedestrian crossing movement, then some parameters inherited from the vehicular parent phase become ineffective by nature, such as yellow and green flash duration. Specially, if the parent vehicular phase has a red/amber, it is replaced by a red time of the same length for the overlapped pedestrian crossing phase; further, if the pedestrian phase has a specified late start time, the red time introduced by the parent phase's red/amber would become ineffective, as the duration of late start is the difference from start time of WALK interval and the start time of the split. If a phase overlaps multiple adjacent parent phases and forms multiple 'connected' occurrences in the cycle, the occurrences of this phase will be merged using the phase merging method in part C.

### *E. Plan Validation*

After plan assembly, the generated plan could be visualized by the signal color sequence for each phase as in Fig. 4, in a VISSIM signal times table style [20].

For a safe plan implementation, validation is conducted on the generated plan. With a conflict matrix, movements with cross conflicts are defined for an intersection. At any time point, for any pair of conflicted movements in the matrix, it is required that at least one of them has red light (except for permissive left-turns and the opposing through movements). Right-turn related merging conflicts, on the other hand, are allowed. In addition, it is also required that pedestrian's walk interval is at least 7 seconds.

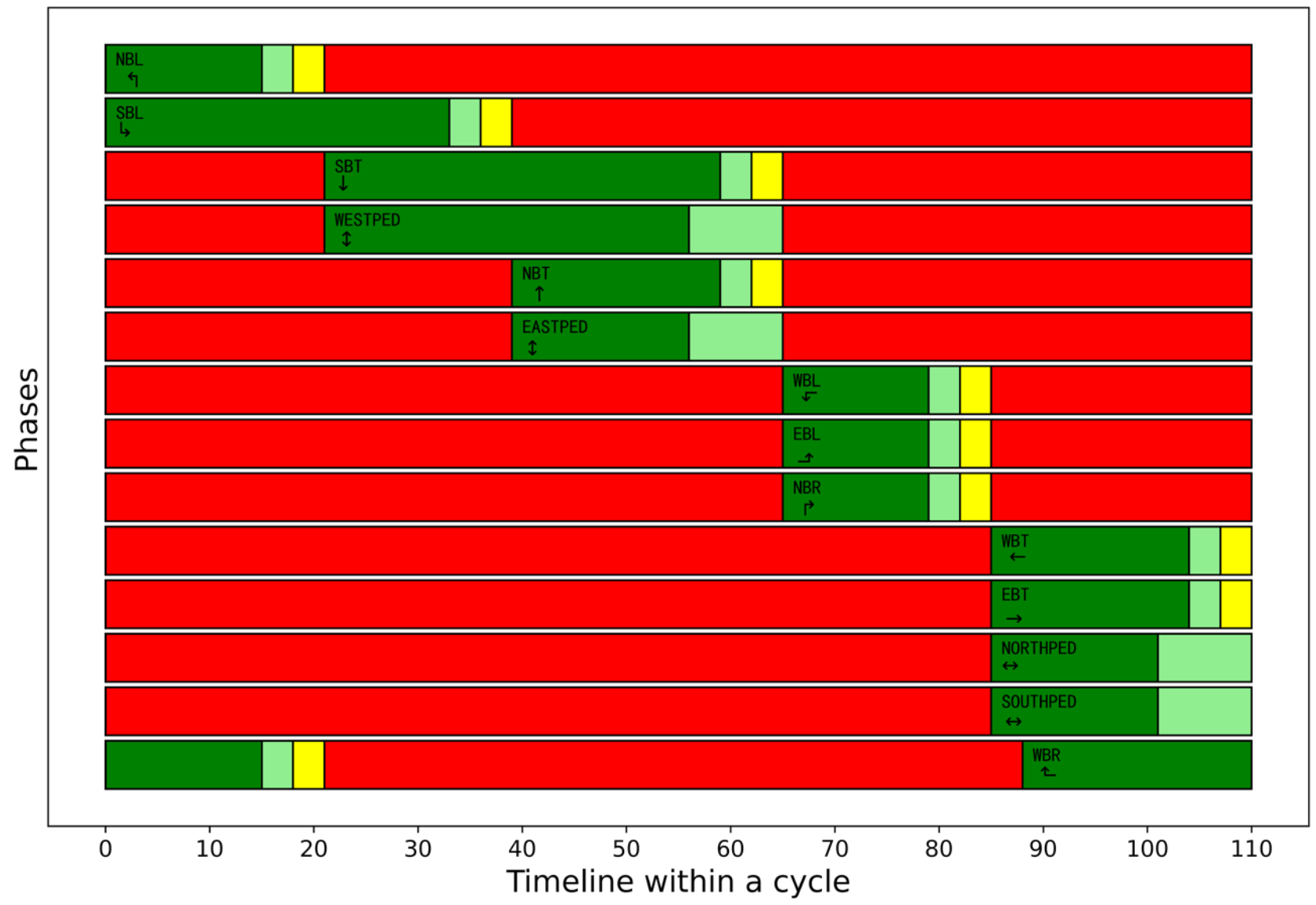

**Fig. 4.** Signal color sequence of each phase.

In this paper, with the hypothetical intersection in Fig. 1, traffic signal lights are assumed to be available for all movements. In practice, however, it is necessary to validate the plan's compatibility with the intersection's signal groups and indications. If the intersection does not support the operation of some phases specified in the description, such as protected right-turns and two-stage pedestrian crossing, users should be reminded to make further modification on the plan. Moreover, traffic signals of any movements not mentioned in the description are treated as non-existent or lights off. Hence, if some critical phases are missing in the plan, such as through movements, users should also be warned.

After plan assembly, if the actual cycle length is not equal to the value in the original plan description, the plan is still valid, as the cycle length may change due to plan editing.

## V. Experimental Results

To validate the proposed method, a test dataset with over 300 plan descriptions in both English and Chinese is created, which covers common plan schemes and description styles. Using four different LLMs, model performance in the accuracy of traffic signal control plan generation is evaluated and analyzed, compared with the 'ground truth' traffic signal color sequences constructed by TSC experts. Based on the results, suggestions are made on the usage and potential applications of the tool.

### *A. Test dataset*

A test dataset with 306 plan descriptions in both English and Chinese is created, including 188 plan creating descriptions and 118 associated plan editing descriptions. The test cases cover common 2 ~ 5 stage plans for four-legged intersections, as well as common phasing schemes for T-intersections and pedestrian crossing intersections. The plan description styles range from professional and precise commands to abbreviated and ambiguous expressions. For each plan description, the traffic light colors of all associated movements are coded second by second in a cycle, as the 'ground truth' for the plan, to enable an automated model performance evaluation. The color and meaning of the codes are listed in Table I.

TABLE I
Color Code of Traffic Lights

| Code | Color and Meaning |
|---|---|
| 0 | Red |
| 1 | Yellow |
| 2 | Green; Pedestrian WALK interval |
| 3 | Green flash; Pedestrian Flashing Don't Walk |
| 4 | Red/amber |
| -1 | Permissive movement; Light off |

To test the model's performance on plan validation, 26 invalid plan descriptions are included in the test dataset, in which conflicted movements are timed simultaneously or walk interval is too short for pedestrian crossings. In the experiments, Chat2SPaT's plan generation result is labeled correct only if it has the same phases and the same traffic light color codes for each second as the ground truth.

### *B. Model Performance*

Four LLMs of different sizes and language backgrounds are tested for the chat-based plan generation process (ChatGPT-4o, Qwen2.5B-72B-Instruct, Deepseek-V3 and Qwen3-32B), with English and Chinese prompts utilized for English and Chinese plan descriptions, respectively.

For each test case, LLMs first process the instructions in the prompts, and then convert the plan description to the formatted phase sequence and phase attribute objects. Based on these results, scripts are used to generate and validate the final signal control plan and compare it with the ground truth traffic light color codes. To account for the randomness in LLMs' reasoning, the plan result is generated and evaluated three times for each case, with a temperature of 0.7 after trial-and-error testing with different parameters.

TABLE II
Accuracy of Plan Generation

| Model | English | Chinese | Overall |
|---|---|---|---|
| ChatGPT-4o | 94.12% | 89.54% | 91.83% |
| Qwen2.5-72B-Instruct | 86.27% | 94.55% | 90.41% |
| Deepseek-V3 | 86.06% | 92.16% | 89.11% |
| Qwen3-32B | 81.82% | 90.20% | 86.06% |

Table II shows Chat2SPaT's performance using four LLMs, with plan generation accuracy shown for English cases, Chinese cases and total. With ChatGPT-4o, the largest LLM among the four, Chat2SPaT achieves the best overall accuracy 91.83%. With a smaller model, Qwen2.5-72B-Instruct, the accuracy remains above 90%. When grouped by languages, an interesting finding is that, all models perform significantly better for tasks when using the language of their own language background, even though all these LLMs are trained to handle multilingual chat and reasoning tasks. In fact, ChatGPT-4o performs the worst for cases in Chinese, even outperformed by the smallest model Qwen3-32B. Yet, ChatGPT-4o and Qwen2.5-72B-Instruct could both achieve an accuracy higher than 94% for English and Chinese plan description cases, respectively.

The experiments show that LLMs could follow the prompts and accurately understand complicated plan structures. For example, in Fig. 5, the plan in test case C11 is described with stage structures nested in a ring structure, and LLMs correctly records the plan structure as instructed in the prompts. Further, the plan description in case C11.1 specifies two permissive right-turn movements in two different styles, and the attributes for both are correctly recorded.

With the flexibility in phase attributes offered by Chat2SPaT, users could generate plans with various special phase treatments. For example, in Fig. 5, the plan in test case C39 specifies two-stage pedestrian crossing for the west and default one-stage pedestrian crossing for the other directions. Further, the plan description in case C39.1 changes SBL to a dummy phase, which is equivalent to removing the associated left-turn signal heads. Such specification could be applied in situations where the intersection channelization changes, such as a time period with variable lanes.

Case C11: 【Plan Description】 First ring WBL 49 s, EBT and SouthPed 53 s, EastPed and NBL 36 s, second ring WestPed 28 s, WBT 74 s.
Case C11.1: EBR lights off from 60 till the end of the cycle. NBR is permissive and runs from 100 to 30.

```
{
  "result1": [{
    "ringStyle": [
      [
        {"WBL": {"split": 49}},
        {"stageStyle": [
            [{"EBT": {"split": 53}}, {"SouthPed": {"split": 53}}]
          ]},
        {"stageStyle": [
            [{"EastPed": {"split": 36}}, {"NBL": {"split": 36}}]
          ]}
      ],
      [
        {"WestPed": {"split": 28}},
        {"WBT": {"split": 74}}
      ]
    ]
  }],
  "result2": [
    {"WBL": {"phaseOrder": 1}},
    {"EBT": {"phaseOrder": 1}},
    {"SouthPed": {"phaseOrder": 1}},
    {"EastPed": {"phaseOrder": 1}},
    {"NBL": {"phaseOrder": 1}},
    {"WestPed": {"phaseOrder": 1}},
    {"WBT": {"phaseOrder": 1}},
    {"EBR": {"phaseOrder": 1, "startTime": 60,
            "endTime": "cycleLength", "isPermissive": 1}},
    {"NBR": {"phaseOrder": 1, "isPermissive": 1,
            "startTime": 100, "endTime": 30}}
  ],
  "result3": null
}
```

Case 11.1

Case C39: 【Plan Description】 NBT and SBT 31 s, NBL and SBL 24 s, EBT and WBT 33 s, EBL and WBL 27 s. Add 2 stage ped crossing for west, add ped phase for north, south and east.
Case C39.1: Change SBL to dummy phase.

```
{
  "result1": [
    {
      "stageStyle": [
        [{"NBT": {"split": 31}}, {"SBT": {"split": 31}}],
        [{"NBL": {"split": 24}}, {"dummyPhase": {"split": 24}}],
        [{"EBT": {"split": 33}}, {"WBT": {"split": 33}}],
        [{"EBL": {"split": 27}}, {"WBL": {"split": 27}}]
      ]
    }
  ],
  "result2": [
    {"NBT": {"phaseOrder": 1}},
    {"SBT": {"phaseOrder": 1}},
    {"NBL": {"phaseOrder": 1}},
    {"dummyPhase": {"phaseOrder": 1}},
    {"EBT": {"phaseOrder": 1}},
    {"WBT": {"phaseOrder": 1}},
    {"EBL": {"phaseOrder": 1}},
    {"WBL": {"phaseOrder": 1}},
    {"WestPedA": {"phaseOrder": 1, "parentPhase": "default"}},
    {"WestPedB": {"phaseOrder": 1, "parentPhase": "default"}},
    {"NorthPed": {"phaseOrder": 1, "parentPhase": "default"}},
    {"SouthPed": {"phaseOrder": 1, "parentPhase": "default"}},
    {"EastPed": {"phaseOrder": 1, "parentPhase": "default"}}
  ],
  "result3": null
}
```

Case 39.1

**Fig. 5.** LLM outputs for test case C11.1 and C39.1.

### *C. Case Study Insights*

In the experiments, various erroneous results in LLM outputs are found, especially when smaller LLMs are utilized. Based on the causes, the errors by LLMs are categorized into three groups.

1) Formatting errors. As introduced in Section III, LLMs are instructed to record the plan structure as a list of json objects labeled as stages or rings, with phases in the same stage or ring written in nested lists. Yet, LLMs may get confused by the usage of json objects and lists, and sometimes omit the label of stage. Common formatting errors could be addressed in the step of data cleansing, as introduced in part A of section IV.

2) Overthinking errors. With a tendency to overthink [21], LLMs sometimes decide to derive redundant results by mathematical calculation, based on available information in the description. However, as LLMs tend to fail in such tasks requiring professional domain knowledge and precise numerical calculations [6], LLMs' calculation results could potentially be erroneous. For example, when the start and end times of a phase are provided, LLMs may go one step further to calculate the duration of the split (or green time), even though LLMs are only instructed to record the specified start and end times. By contrast, it is observed that when the split duration is provided by users, LLMs would not infer the start and end times for the phase, as their connections are implicitly hidden in the stage or ring based plan structures.

In LLM outputs, it is difficult to distinguish user specified values from LLMs' overthinking results. Therefore, prompts and fault tolerance functions are designed and iteratively refined, to reduce the chance and impact of LLMs' typical overthinking errors. For the example above, when start time, end time and split duration of a phase are all recorded in LLM outputs, Chat2SPaT would ignore the split duration, and locate the phase in the cycle with only start and end times specified by the user.

3) Semantic comprehension errors. Conforming to LLM's scaling law [22], the performance of Chat2SPaT declines when using Qwen3-32B, with an overall accuracy 3~6% lower than the larger models. In particular, for English cases, the accuracy drops by 12.3%, compared with ChatGPT-4o. As many formatting errors and overthinking errors are addressed by fault tolerance functions, the accuracy drop could be mostly attributed to LLMs' semantic comprehension errors. Smaller LLMs have more difficulty understanding the plan structure in the descriptions, and may get confused dealing with phase attributes, such as parent phase matching and phase order counting.

Hence, to improve the accuracy of Chat2SPaT, it is recommended that researchers and traffic practitioners select appropriate LLMs based on the results in Table II. Using fine-tuned LLMs trained for TSC tasks, Chat2SPaT's performance could potentially be further improved. To reduce the chance of LLM hallucination, it is suggested to use simple and consistent plan descriptions, in a clear keyword and value style.

The experiment results validate the proposed method for chat-based plan generation, and demonstrate Chat2SPaT's application potential in the practice of traffic signal control plan management. Chat2SPaT could be integrated into traffic signal control software or microscopic traffic simulation software (such as SUMO), to reduce traffic engineers' repetitive manual parameter inputting for plan creation and modification. Based on the proposed pipeline, the open-sourced prompts and plan assembly algorithm could be further updated, to allow for customized implementation of the tool. For example, the available signal heads and movements of an intersection could be used as an additional input, so that Chat2SPaT could conduct a more informed plan generation and validation.

## VI. Summary and Conclusions

Aiming at boosting efficiency for traffic agencies, the paper proposes an LLM-based tool, Chat2SPaT, to enable the management of traffic signal control plan with chats. The model leverages LLMs' capability to understand users' semi-structured and ambiguous descriptions on plan creating and editing, records plan structure and phase attributes, and finally assembles the complete plan with scripts. With curated prompts, the model breaks down the complexity and variations of traffic signal control plans, and instructs LLMs to record phase attributes as simple key-value information, without requirement for further reasoning or math calculation to derive temporal relationship among phases. This design makes the task easier for LLMs, and thus reduces chance of hallucination and improves the model's robustness against LLM incapability. With minimal output tokens required, the response time of LLMs is also reduced.

A test dataset with over 300 plan descriptions is created, covering common plan schemes and description styles. Experiments show that ChatGPT-4o based Chat2SPaT can generate plans with an overall accuracy of 91.83% and an accuracy of 94.12% for English cases. Using smaller LLMs such as Qwen2.5-72B-Instruct, the overall plan generation accuracy remains above 90%, with an accuracy of 94.55% for Chinese cases.

In future works, Chat2SPaT could be integrated into signal control platforms or traffic simulation software. More functionalities should be developed to fully leverage the practical value of the tool, including the management of day-of-week plan / time-of-day plan, and the support on more transportation modes (such as bike, equestrian and bus). Besides, in addition to pre-timed controls, the proposed method could also be extended to serve dynamic controls, such as specifying detectors and adjusting control parameters for actuated controls. As the first benchmark to extensively evaluate LLMs' capability of understanding traffic signal control plans, the proposed method bridges the gap between semi-structured plan descriptions and structured SPaT data, serving as a potential new building block for a more accurate and versatile application of LLMs in the field of ITS.